\documentclass[runningheads]{llncs}

\usepackage{hyperref}      
\usepackage[all]{hypcap} 

\usepackage{graphicx}

\usepackage{amsmath,amssymb}
\usepackage{bm}
\usepackage{mathdots}

\usepackage{xcolor}
\definecolor{mygreen}{RGB}{55,125,34}

\usepackage{listings}
\usepackage{tikz}
\usetikzlibrary{calc}

\usepackage{booktabs}
\usepackage{multirow}

\usepackage{algorithm}
\usepackage{algpseudocode}

\begin{document}
\title{New methods for metastimuli: architecture, embeddings, and neural network optimization}
\titlerunning{Metastimuli: architecture, embeddings, ANN optimization}
%
\author{Rico A.\,R.\ Picone\inst{1,2}\orcidID{0000-0002-5091-5175} \and
Dane Webb\inst{1}\orcidID{0000-0003-3722-617X} \and
Finbarr Obierefu\inst{3}\orcidID{0000-0001-8935-7222} \and
Jotham Lentz\inst{2}\orcidID{0000-0003-1190-1632}}
\authorrunning{R.\,A.\,R.\ Picone et al.}
%
\institute{Saint Martin's University, Lacey WA 98503, USA 
\email{rpicone@stmartin.edu}\\
\url{http://www.stmartin.edu}\and
Dialectica LLC, Olympia WA 98501, USA
\email{rico@dialectica.io}\\
\url{http://dialectica.io}\and
Universit\'e Bourgogne Franche-Comt\'e, Besan\c{c}on, France\\
\url{http://www.ubfc.fr}}
\maketitle              
\begin{abstract}
Six significant new methodological developments of the previously-presented ``metastimuli architecture'' for human learning through machine learning of spatially correlated structural position within a user's personal information management system (PIMS), providing the basis for haptic metastimuli, are presented.
These include architectural innovation, recurrent (RNN) artificial neural network (ANN) application, a variety of atom embedding techniques (including a novel technique we call ``$\nabla$'' embedding inspired by linguistics), ANN hyper-parameter (one that affects the network but is not trained, e.g.\ the learning rate) optimization, and meta-parameter (one that determines the system performance but is not trained and not a hyper-parameter, e.g.\ the atom embedding technique) optimization for exploring the large design space.
A technique for using the system for automatic atom categorization in a user's PIMS is outlined.
ANN training and hyper- and meta-parameter optimization results are presented and discussed in service of methodological recommendations.

\keywords{Design: Human Centered Design and User Centered Design \and Design: Information design \and Technology: Augmented Reality and Environments \and Technology: Haptic user interface \and Technology: Intelligent and agent systems \and Technology: Natural user interfaces (NUI).}
\end{abstract}
%
%
%

\section{Objective and significance}

\begin{figure}[bth]
\centering
\includegraphics[width=.6\linewidth]{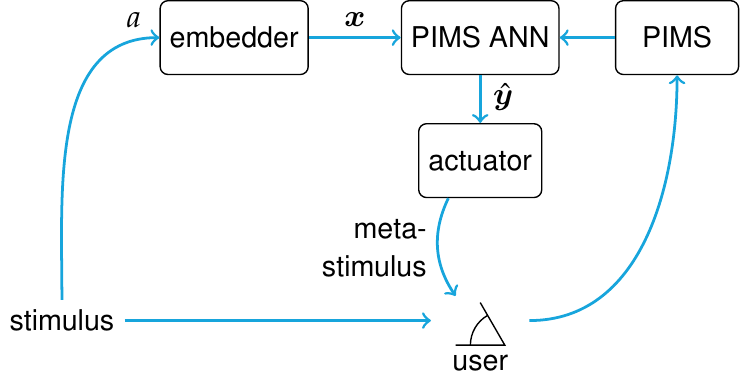}
\caption{The metastimuli system architecture. An atom $a$ of stimulus, experienced directly by the user, is, in textual form, also directed through an embedder block that embeds the atom as a vector $\bm{x}$, which is the input to an artificial neural network (ANN) pre-trained on the user's structured personal information management system (PIMS). The output of the ANN $\bm{\hat{y}}$ is a low-dimension real vector representing the structural ``location'' of the atom $a$, which is converted by an actuator into a metastimulus (e.g.\ haptic stimulus). \cite{picone_2021}}
\label{fig:metastimuli_architecture}
\end{figure}

We present six significant developments in the metastimuli architecture introduced by \cite{picone2020}.
The goal of the architecture remains: to improve human learning of textual source material (i.e.\ text, audio, video with dialog) by presenting a user with, in addition to their direct experience of the material, correlated \emph{metastimuli} that represent (through time) the structural ``position'' the source material has in their own personal information management system (PIMS).
Some of the methods of the original architecture \cite{picone2020} are augmented in the architecture presented here, illustrated in \autoref{fig:metastimuli_architecture}.
In particular, the \emph{classification} of atoms of information, which was a discrete process in the original architecture, is here integrated into the training of the primary artificial neural network (ANN) of the system.
In fact, explicit classification has been circumvented altogether.
Furthermore, the original method of \emph{atom embedding}, which is similar to sentence embedding with its basis in word embedding, is compared to several newer techniques.
Two variations of PIMS-trained artificial neural network (ANN) are compared: a feedforward is compared with a recurrent neural network (RNN).
Finally, the atom embedder and PIMS-trained ANN can be straightforwardly adopted as a standalone \emph{classifier} in its own right; however, since explicit classification is not strictly required for metastimuli, its presentation is limited to a structural exposition.

\section{Methods}

The six methodological innovations are presented below.

\subsection{Integrated PIMS classification}

\begin{figure}
\centering
\includegraphics[width=.6\linewidth]{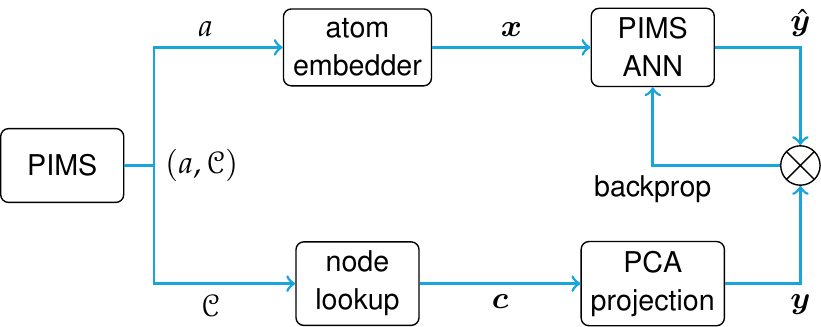}
\caption{Training of the PIMS ANN. An atom $a$ with category label $\mathcal{C}$ (corresponding to the structural ``location'' of $a$) from a training set of the user's PIMS is directed to two paths, the upper for the atom and the lower for the category label. The atom embedder encodes the atom $a$ as vector $\bm{x}$, which is the input to the PIMS ANN. The category $\mathcal{C}$ is first given a high-dimensional one-hot vector representation $\bm{c}$, then projected via principal component analysis (PCA) projection to low-dimensional $\bm{y}$, which is the numerical label of the atom. The difference between this and the ANN output $\bm{\hat{y}}$ is then backpropagated. \cite{picone_2021b}}
\label{fig:training_loop}
\end{figure}

\autoref{fig:training_loop} illustrates the new training loop for the PIMS ANN.
A key innovation here is that the projected PIMS representation is trained-into the ANN such that a separate classification process is circumvented altogether.

\subsection{Recurrent PIMS-trained ANN}

We compare the performance of a FFNN and a RNN for the PIMS ANN object in \autoref{fig:training_loop} and \autoref{fig:atom_classifier}. 
FFNNs are a simple form of neural networks with a lot of general utility.
Connections in FFNNs do not loop and the information always moves ``forward'', (input layer to hidden layer to output layer).
RNNs are excellent for natural language processing tasks. 
RNNs are able to use the sequential nature of natural language in the neural network architecture. 

\subsection{Null set validation}
\label{sub:nullset}
One of the conventionally presented results of an ANN training is a plot of training and (sometimes) testing loss versus training epoch.
A downward trend of training loss signifies learning, whereas the downward trend of testing loss signifies the generality of that learning beyond the training set.
Overfitting occurs when the ANN learns to predict the testing data to the detriment of generality therebeyond.
In the context of natural language processing (NLP), overfitting is the learning of specific semantic (meaning) constructions to the detriment of recognizing similar constructions, observable when the training loss continues to decrease while the testing loss begins to increase.
Mitigation techniques for overfitting include the familiar ``dropout'' method in which certain data is ignored during parts of training.

\begin{figure}
    \centering
    \includegraphics[width=.8\linewidth]{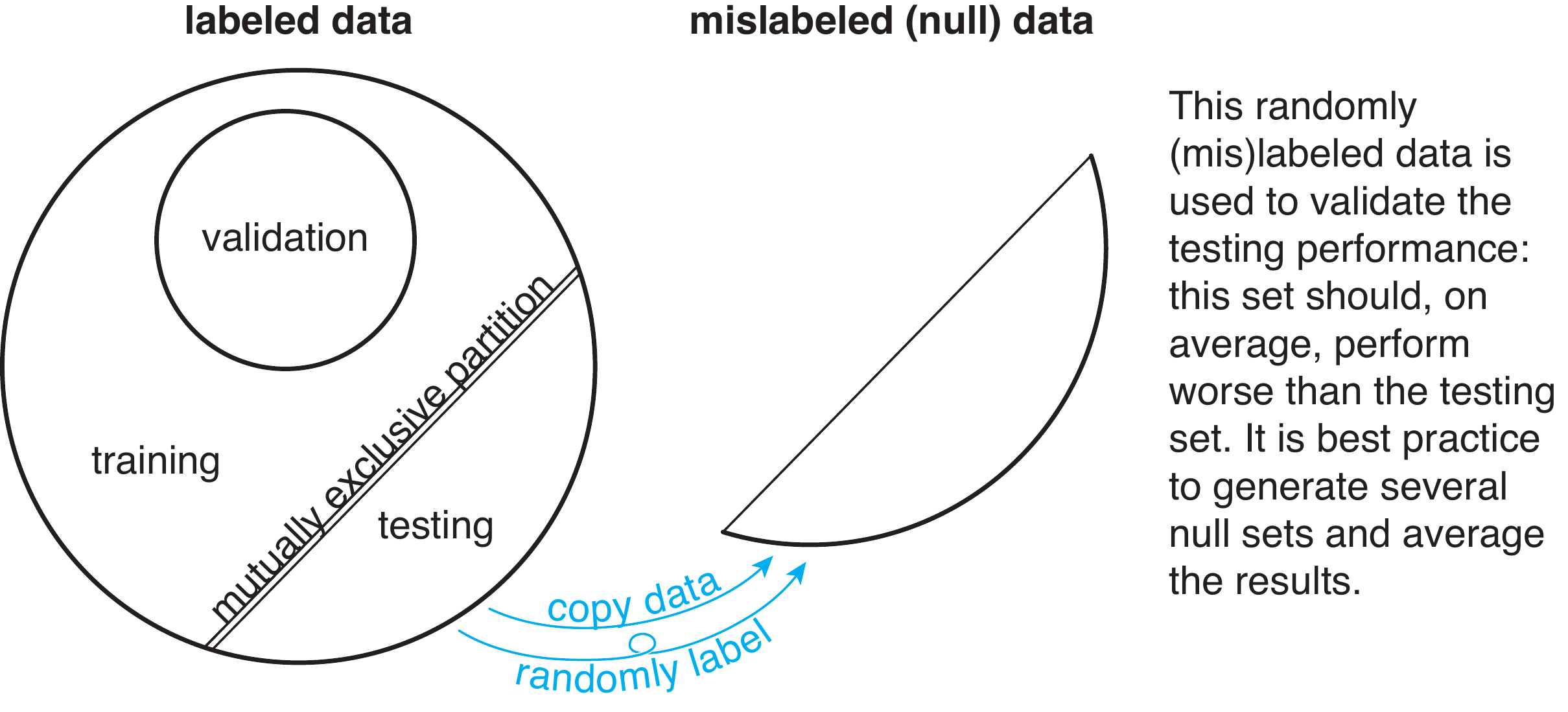}
    \caption{An illustration of the data partitions, including the null partition, a copy of the testing data labeled randomly. This technique is derived from the experimental and statistical validation method of the same name. Despite the significant computational overhead, computing the loss for several null sets and averaging them for comparison to training loss (which should be lower) is recommended at least during early development of a new model. \cite{picone_2020}}
    \label{fig:null_dataset}
\end{figure}

A question remains, however: how significant is the learning, really?
The magnitude of loss is not easily compared between data sets and learning methods, and there is no readily available ``baseline'' for comparison.
Now consider an ANN's prediction performance on a set of testing data that is randomly labeled, meaning its semantic content is as unrelated to its labels as possible.
Such a testing set should show, on average, no learning (reduced loss) and can be thought of as representing a sort of contextually relevant ``random guess'' of predictions.
We call this data a \emph{null set}; this is illustrated in \autoref{fig:null_dataset}.

It is difficult in practice to construct a null set that is truly unrelated to the content, however, given that the content and labels, despite their random assignment, are always already somewhat related.
Therefore, a small amount of learning does in fact occur, but it represents a contextual ``baseline'' for training and testing loss.
The null loss baseline places a lower-bound on performance metrics.

It is best to generate several randomly labeled copies of the testing data, test the network on all of them and averaging the loss (to mitigate any ``luck'' in the randomly assigned labels).
This does increase computational cost, especially if undertaken at every training epoch, so its use should be limited to early development and final testing of an ANN model.

The term ``null'' is used to draw attention to the similarity of this technique with the \emph{null hypothesis testing} of statistics and ``null sample'' measurement in experimental physics, in which the results of a challenging experiment are considered well-validated when the only difference between two measurements is the sample, one of which is of interest and the other of which yields a baseline, zero, or null result.

\subsection{Atom embeddings}

Word embeddings are representations of words in a real vector space that maintains word context and relations thereamong. 
Embeddings can be learned via a neural network from training data that includes many examples of natural-language usage of the words in question \cite{charniak2018}. 
Embeddings of words that, in the training set, appear in similar contexts, drift toward each other over the course of training. 

Sentence embeddings are similar to word embeddings in that they encode a sentence as a vector that represents the semantics of each sentence. 
Sentence embeddings are typically mapped or learned from component word embeddings. 
Good embeddings filter some of the sharper contrasts in sentences and help convey the context, intention, and subtleties in the text. 

Sentence embedding techniques can be adapted to word sequences of different length.
Our application calls for approximately paragraph-length sequences we call \emph{atoms}.

\subsubsection{Embedding atoms}
\label{sec:embedding_atoms}

In both the training loop, \autoref{fig:training_loop}, and in application, \autoref{fig:atom_classifier}, the atom embedder converts atoms to vector representations that the PIMS ANN can process. 
The atom embedder contains a word embedder that encodes each word into its word embedding vector.
An atom $a$ of word embedding vectors is then encoded into an atom embedding $\bm{x}$.

Two word embedders are evaluated. 
The first is trained with a relatively small custom corpus and the second is trained with a large publicly available corpus.
The custom corpus contains a mechatronics textbook and a dynamic systems textbook written by one of the authors. 
The large dataset is from TensorFlow Datasets, the ``scientific-papers'' corpus \cite{Cohan_2018}, which contains over 200,000 scientific papers from \url{https://ArXiv.org}.

Quality of embeddings depends on the specificity of the dataset to the use-case and on the size of the dataset. 
We compare the performance of the small but specific custom corpus and the large but less-specific corpus.

There are several sentence embedding methods that can be applied to atoms. 
We evaluate the relative performance of several of these methods. 

Four candidate methods are described below.

\begin{description}
    \item [Bag of Words (BOW)] sums or averages component word embedding vectors. 
    While surprisingly accurate for its simplicity, word order, distance between words, and semantics are lost. \cite{le2014distributed}
    \item [Distributed Memory ``Paragraph'' Vector (PVDM)] assigns each sentence (``paragraph'') in an atom a trained identification (ID) vector. 
    This ID vector and its sentence's word vectors are trained together. \cite{le2014distributed} 
    \item [Smooth Inverse Frequency (SIF)] computes sentence embeddings as a weighted average of word vectors. 
    \cite{arora2016simple}
    \item [Universal Sentence Encoders] Averages all the words of a sentence before feeding into multi-layered ANN.
    \cite{cer2018universal}
\end{description}

The results of \autoref{sec:results} include the first two above, and the novel $\nabla$-embeddings presented below.

\subsubsection{$\nabla$ embeddings}

The father of linguistics, Ferdinand de Saussure introduced the concept of the \emph{differential} value of meanings and words.
First considering the signifier or word.
\begin{quote}
The important thing in the word is not the sound alone but the phonic differences that make it possible to distinguish this word from all others, for differences carry signification.
This may seem surprising, but how indeed could the reverse be possible? Since one vocal image is no better suited than the next for what it is commissioned to express, it is evident, even \emph{a priori}, that a segment of language can never in the final analysis be based on anything except its noncoincidence with the rest. \emph{Arbitrary} and \emph{differential} are two correlative qualities. \cite[p.~118]{saussure1916}
\end{quote}
So the ``vocal image'' or word is itself \emph{arbitrary} and only takes on meaning in its \emph{difference} from others.
Consider the following with regard to meaning.
\begin{quote}
Instead of pre-existing ideas then, we find [...] values emanating from the system. When they are said to correspond to concepts, it is understood that the concepts are purely differential and defined not by their positive content but negatively by their relations with the other terms of the system. Their most precise characteristic is in being what the others are not. \cite[p.~117]{saussure1916}
\end{quote}
This differentiality holds for both the word and its meaning.

Word embeddings differentially (usually via recurrence) embed word meanings.
Therefore, when considering a sentence or paragraph embedding---for us, an \emph{atom} embedding---it is worth considering differentiality.
For instance, consider the sentence \emph{She opens tonight.}
Not until the final word do the preceding words take on their proper meaning: \emph{She opens \dots{}} could go a different way, such as \emph{She opens presents.} So \emph{tonight} actually fixes the meaning of \emph{opens}, in this case in the sense: she is the opening act.

There is much nuance, here.
The bag-of-words methods of sentence embedding that sum or average the word embeddings in a sentence retain some differentiality of meaning left over from the word embeddings, but they ignore \emph{order}, which impacts the differential meanings.
Several newer methods, including those listed above, encode in some way the ordering of words, but increase (system and computational) complexity significantly.

We introduce a simple approach we call \emph{$\nabla$-embeddings}.
Consider atom $a$ and the difference between two sequential word embedding vectors $\bm{e}_i$ and $\bm{e}_{i+1}$,
\begin{align}
\nabla_i^1 = \bm{e}_{i+1} - \bm{e}_i.
\end{align}
Furthermore, consider the next-level of difference of differences
\begin{align}
\nabla_i^2 = \nabla_{i+1}^2 - \nabla_i^2
\end{align}
such that $\nabla_i^j$ signifies the $i$th difference at the $j$th level.
Proceeding in an array over an atom with $\nu$ words, we obtain the following inverted Pascal's triangle, standing on its head if you will.
{\renewcommand{\arraystretch}{1.75}
\begin{align*}
\begin{array}{ccccccccccccc}
\bm{e}_1\phantom{xx} & & \bm{e}_2\phantom{x} & & \bm{e}_3 & & \cdots & & \bm{e}_{\nu-2} & & \bm{e}_{\nu-1} & & \bm{e}_{\nu} \\
 & \nabla_{1\phantom{+ 2}}^1 & & \nabla_{2\phantom{+ 2}}^1 & & & \cdots & & & \nabla_{\nu-2}^1 & & \nabla_{\nu-1}^1 &\\
 & & \nabla_{1\phantom{+ 2}}^2 & & & & \cdots & & & & \nabla_{\nu-1}^2 & & \\
 & & & \ddots\phantom{xx} & & & \vdots & & & \iddots & & & \\
 & & & & \ddots & & \vdots & & \iddots & & & &  \\
 & & & & & & \phantom{xx}\nabla_1^{\nu-1} & & & & & & 
\end{array}
\end{align*}
}

Furthermore, consider a sum $\bm{x}^j$ over a level $j$,
\begin{equation}
    \label{eq:multi-level-nabla}
    \bm{x}^j = \sum_i \nabla_i^j.
\end{equation}
This is a \emph{$\nabla^j$-embedding} of the atom $a$.
Note that even a $\nabla^1$-embedding retains some differential information.

This new embedding method is applicable to atoms, paragraphs, and sentences.
It is included among the established atom embedding methods in the results of \autoref{sec:results}.

\subsubsection{Keyword weighting} 
Artificial neural networks encode a large range of subtle and crude characteristics/representation of text. 
We expect that much correct categorization could be achieved with simple keyword\footnote{Using available machine learning natural language libraries, such as the popular NLTK used here \cite{BirdKleinLoper09}, it is straightforward to lemmatize a category name in order to generate relevant keywords automatically.} identification (e.g.\ the category \emph{voltage} likely applies to an atom containing the word \emph{volts}), so the ANN should include similar functionality.

This functionality is trained into the network by multiplying keyword vectors by a keyword weighting factor that scales the corresponding embedding vectors inflate atom vectors, effectively expressing the greater importance of keywords in an atom embedding.

A concern with weighting the keywords is that the ANN may ignore semantic content and rely too much on simple keyword recognition.
Dropout that simply removes random atoms from each training cycle, already used to reduce general overfitting, is one method to counter this concern.
Another is to include the keyword scaling factor as meta-parameter to be optimized, as described below.
In this way, if semantic content is undervalued, the optimizer will adjust accordingly.


\subsection{Meta- and hyper-parameter optimization}


Optimization weights and biases of a ANN, are commonly optimized by gradient descent algorithms.
Gradient descent algorithms are fast and are guaranteed to reach a minimum. 
ANN training space is not plagued by large numbers of local minimum. 
Instead, vanishing gradients and saddle points are issues.

Alternatively, the meta-parameters and the remaining hyper-parameters do present a plane with many local minima.
The hyper-parameters are tuned using keras-tuner.
Keras-tuner includes three, not including sklearn, derivative-free optimizers.
Random search, bayesian, and hyperband are included within the keras-tuner module.

Meta-parameters are optimized with a pattern search algorithm.
Any derivative free algorithm could be substituted that works well with integer objective variables, scales search range near minimum, and is quick.

{\renewcommand{\arraystretch}{1.3}
\setlength{\tabcolsep}{1ex}
\begin{table}[]
    \begin{center}
    \caption{A list of hyper-parameters optimized via keras-tuner and their possible values. The hyparameters are partitioned into those related to the ANN architecture and those related to the ANN training. }
    \label{tab:hyper_optim}
    \begin{tabular}{lll}
        \toprule
        & hyper-parameter & possible values \\
        \midrule
        \multirow{4}{*}{ANN architecture}
        & weights/biases & $\mathbb{R}$ \\
        & activation function & $\tanh$, $\sigma$, $S$ \\
        & hidden layers & $\mathbb{Z}_+$ \\
        & features & $\mathbb{Z}_+$ \\ \cline{2-3}
        \multirow{3}{*}{ANN training}
        & gradient-based optimizer parameters & optimizer-dependent \\
        & learning rate & $\mathbb{R}_+$ \\
        & training epochs & $\mathbb{Z}_+$ \\
        \bottomrule
    \end{tabular}
    \end{center}
\end{table}
}

\subsubsection{Hyper-parameter optimization}
Hyper-parameter optimization is handled by the Keras-tuner library.
The three tuners available in the library are random search, bayesian optimization, and hyperband.
For an explanation of the random search algorithm see Rastrigin \cite{rastrigin1963convergence}, for bayesian optimization see Pelikan et al.\ \cite{pelikan1999boa}, and for hyperband see Li et al.\ \cite{li2017hyperband}.
The parameters optimized by the hyper-parameter optimization are shown in \autoref{tab:hyper_optim}.

\subsubsection{Meta-parameter optimization}
Pattern search is one of many derivative-free optimization algorithms that be acceptable.
Pattern search has all the traits required.
Pattern search is used because it is an algorithm that is relatively easy to implement.

Further optimization could be conducted by comparing additional search algorithms.
Choosing the best meta-parameter optimization is a "trial and error" search.
Finding the optimal meta-parameter optimizer is too resource expensive at this stage.

Algorithm \ref{al:pattern_search_algorithm} details the pattern search algorithm where $X^{(\text{new})}$ is the new base point, $X'$ is the new temporary base point, $X$ is the initial base point, $\delta$ is the pattern step size, and $\alpha$ is an acceleration factor.


\begin{algorithm}
    \caption{Pattern search algorithm \cite[p.~51]{bozorg2017meta}} 
    \label{al:pattern_search_algorithm}
    \begin{algorithmic}
        \Procedure{PatternSearch}{$input$,$variables$,$here$}
        \State Define parameters for the algorithm.
        \State Generate a random base point $X$.
        \While{The difference between the previous base point best fitness and the current base point best fitness is greater than the minimum difference.}
            \State Generate exploratory points in a pattern, the mesh, around the base point $X$.
            \State $X^{(\text{new})}=$ the best exploratory point
            \If{The fitness of the best exploratory point $X^{(\text{new})}$ is better than the fitness of the base point $X$}
                \State $X^{(\text{new})} \leftarrow \text{best exploratory point}$
                \State $\mu \leftarrow \mu_0$ \Comment{reset mesh size}
                \While{$X^{(\text{new})}$ is better than $X$}
                    \State $X' \leftarrow X$ \Comment{set exploratory point as new base point}
                    \State $X \leftarrow X^{(\text{new})}$
                    \State $X^{(\text{new})} \leftarrow X' + \alpha \cdot (X - X')$ \Comment{pattern move}
                    \State Generate exploratory points in a pattern around the new base point.
                    \State $X^{(\text{new})} \leftarrow $ the best new generated point
                \EndWhile
            \Else{
                $\mu^{(\text{new})} \leftarrow \mu - \delta$ \Comment{decrease the mesh size}
            }
            \EndIf
        \EndWhile
        \EndProcedure
    \end{algorithmic}
\end{algorithm}

{\renewcommand{\arraystretch}{1.3}
\setlength{\tabcolsep}{1ex}
\begin{table}[]
    \begin{center}
    \caption{A list of meta-parameters optimized through pattern search and their possible values. The meta-parameters are partitioned into those related to the atom embeddings and those related to the ANN.}
    \label{tab:meta_optim}
    \begin{tabular}{lll}
        \toprule
        & meta-parameter & possible values \\
        \midrule
        \multirow{4}{*}{embeddings}
        & projection/output dimensions & $\mathbb{Z}_+$ \\
        & word em.\ model/input dimensions & $\mathbb{Z}_+$ \\
        & keyword weighting & $\mathbb{R}_+$ \\
        & atom embedding method & BOW$\Sigma$, , BOW$\mu$, PVDM, $\nabla$ \\ \cline{2-3}
        \multirow{6}{*}{ANN}
        & gradient-based optimizer & SGD, Adam, AdaGrad, \\
        & & AdaDelta, AdaMax, \\
        & & RMSprop \\
        & ANN architecture & FFNN, RNN \\
        & tuner optimizer & random search, hyperband, \\
        & & Bayesian \\
        & tuner parameters & optimizer-dependent \\
        \bottomrule
    \end{tabular}
    \end{center}
\end{table}
}


\begin{figure}
\centering
\includegraphics[width=.7\linewidth]{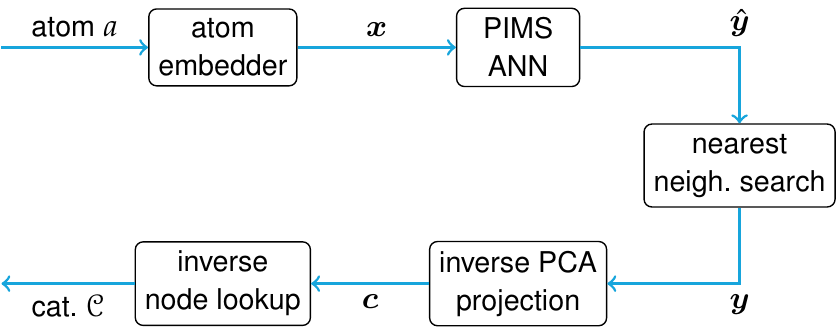}
\caption{An atom classifier derived from the atom embedder, PIMS ANN, the inverse PCA projection, and an inverse node lookup. The ANN has been trained on the user's PIMS structure, thereby making its output a low-dimensional estimate of the atom's ``location'' in the user's PIMS. Converting this estimate into ``node space'' requires a nearest-neighbor search, which can occur as shown, before inverse projection, or after. \cite{picone_2021c}}
\label{fig:atom_classifier}
\end{figure}

\subsection{PIMS ANN as a classifier, a bonus application}


As has been noted, the new metastimuli architecture circumvents the need to explicitly classify each atom.
However, and this classification is in a sense implicit in the output of the PIMS ANN $\bm{\hat{y}}$.
All that is needed to yield an explicit classification of each atom is a relatively simple procedure depicted in \autoref{fig:atom_classifier}: a (one) nearest-neighbor search, an inverse principle component analysis (PCA) projection, and an inverse node lookup.
This is superfluous for metastimuli, but can be deployed as an auto-classifier for other applications with similar PIMS structures.

\section{Results}
\label{sec:results}

Several new methods of metastimuli generation are presented, including an improvement to the overall metastimuli architecture.
Variations on each method are compared, resulting in application-specific recommendations.
The software tools we develop and present are made available as public repositories (those already available are \cite{rico_picone_2020_3633355,danewebb_class,danewebb_tag}).

\subsection{Learning by epoch}

During optimization, each generated ANN model with its meta-parameter and hyper-parameter set is trained for 10 epochs.
This is not enough training to produce a usable model but is sufficient as a comparison.
Better meta- and hyper-parameter sets produce better models that demonstrate lower fitness.
The short training is necessary to conserve resources as a single optimization may require hundreds of models to be trained.

Therefore, at the end of optimization, the model with the optimal meta- and hyper-parameters must be trained significantly longer to complete the optimization process with a fully optimized ANN.

In deep learning, which we use here, longer training will continue to improve the loss of the ANN for the dataset upon which it trains.

The optimal ANN with meta- and hyper-parameter is trained for 200 epochs.
After each training epoch, the ANN evaluates the testing dataset, which is labeled data withheld during the training process.
Evaluating the testing set after each training epoch demonstrates the ANN is learning from similar data and not merely memorizing the dataset it trains on, which is called overfitting.

Upon completion of the test dataset evaluation, the model was tested on the null dataset. 
For a full discussion on the use of the null dataset, see subsection \ref{sub:nullset}.

\begin{figure}[]
\centering
    \begin{tabular}{l}
    \includegraphics[width=.9\linewidth]{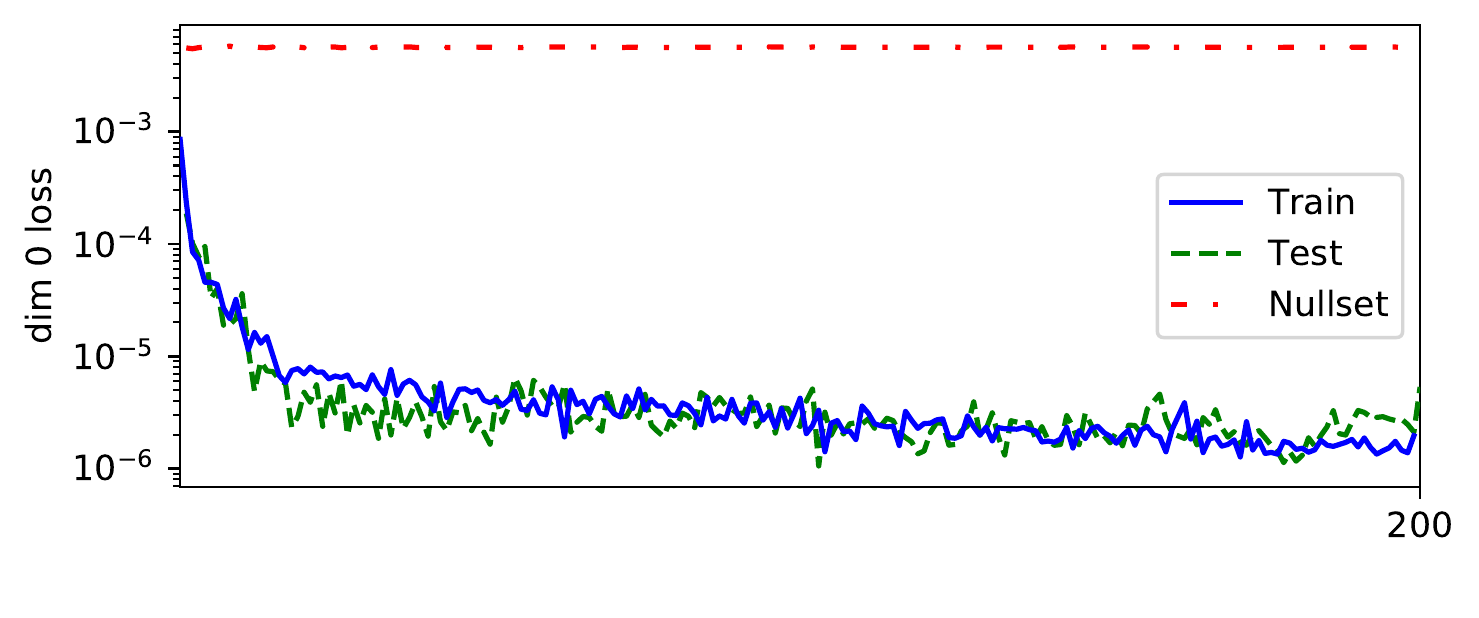} \\
    \includegraphics[width=.9\linewidth]{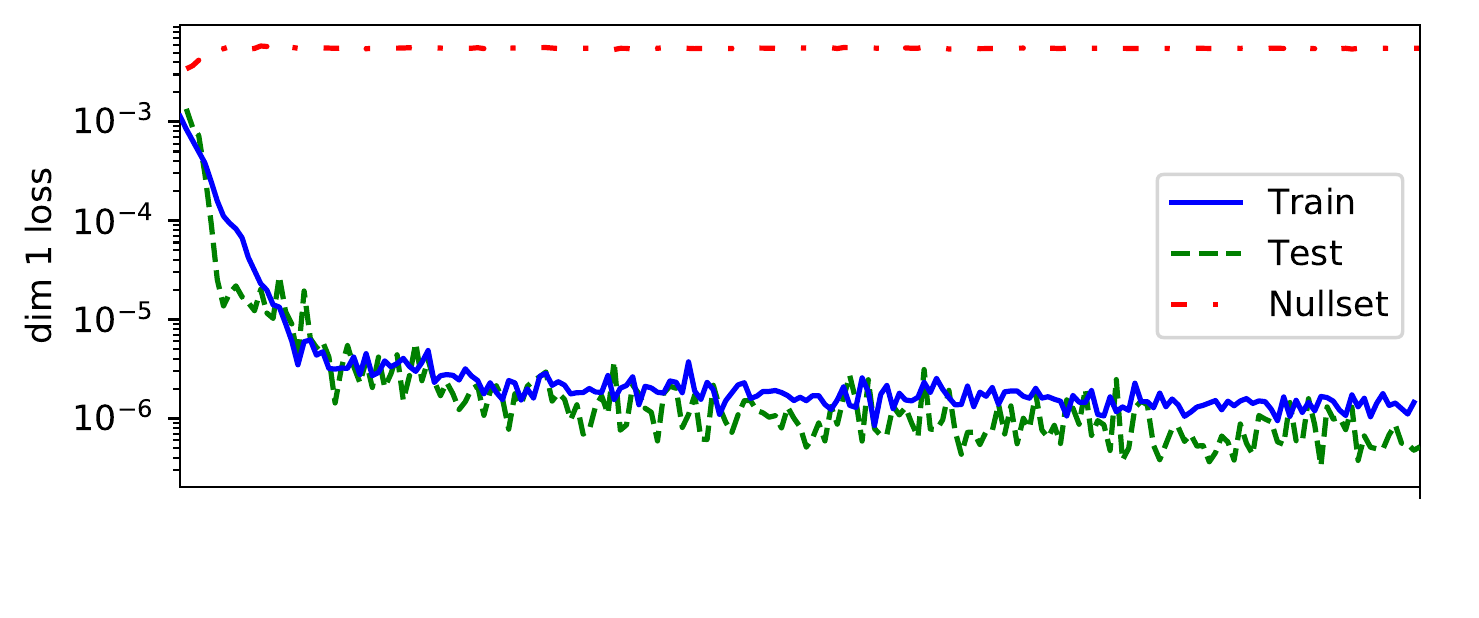} \\
    \includegraphics[width=.9\linewidth]{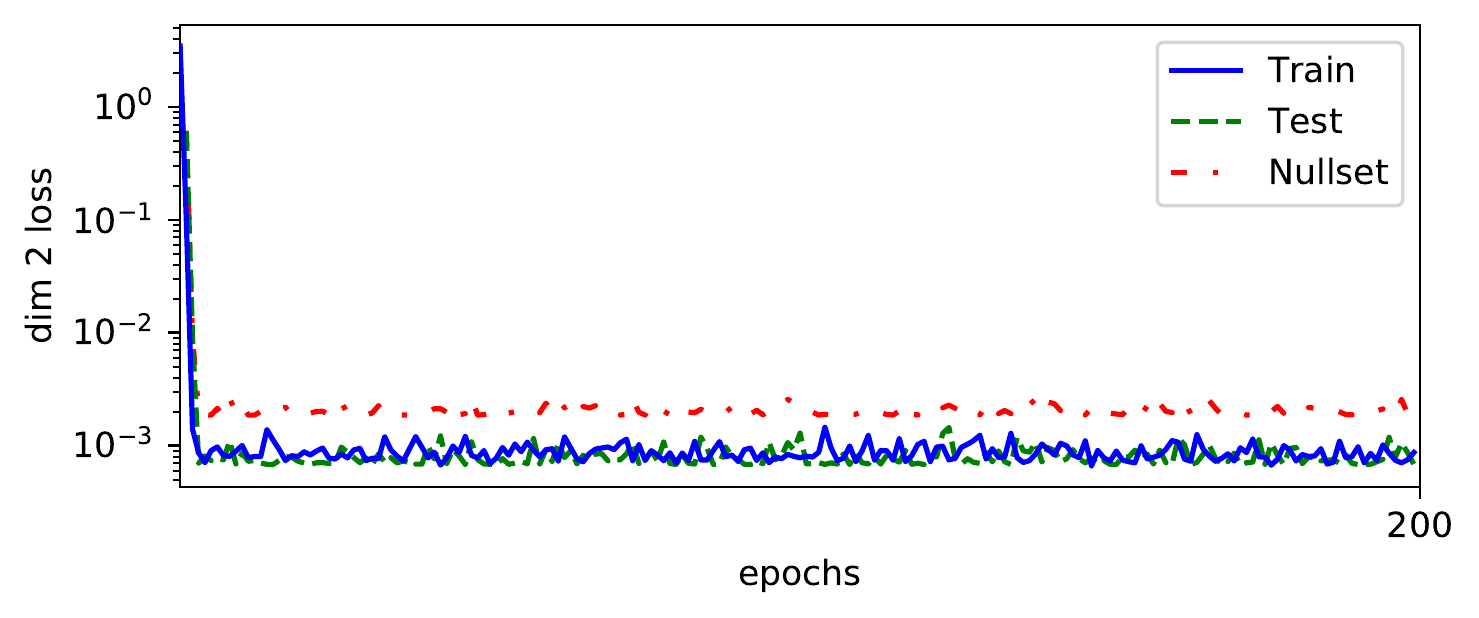} \\
    \includegraphics[width=.9\linewidth]{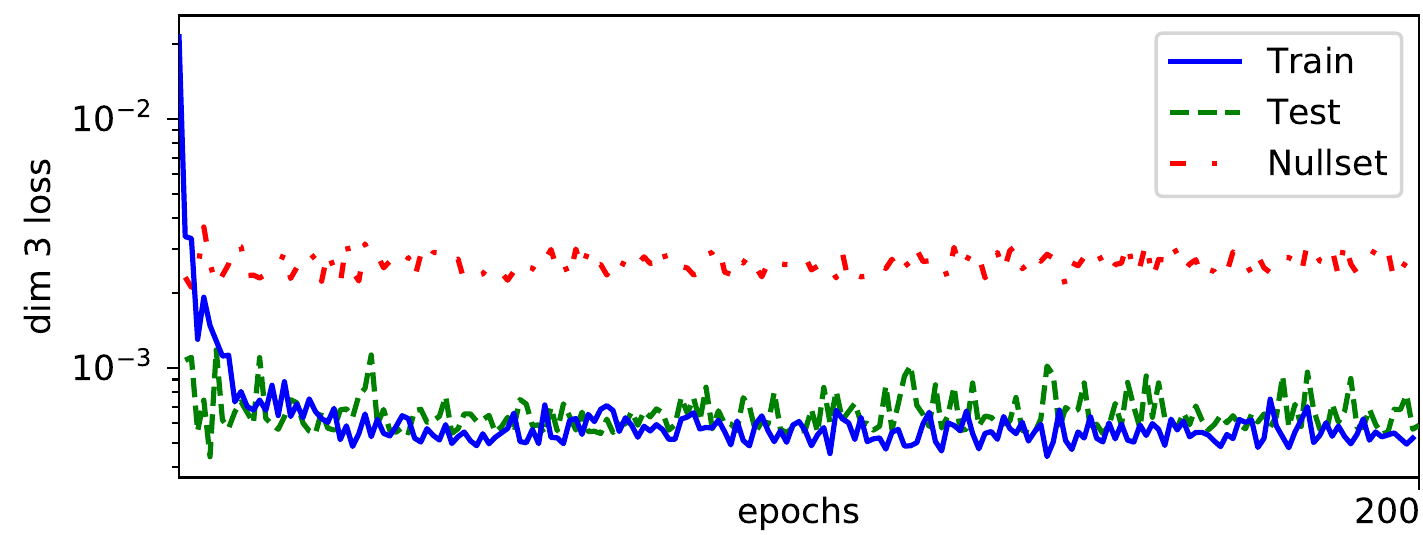}
    \end{tabular}
\caption{The loss for training, testing, and null data by training epoch with the optimal meta- and hyper-parameters. As we expect, the null loss, which is the mean loss of five models, performs poorly compared to the training and testing loss. The training loss is used for ANN backpropagation. The testing loss shows how the model performs at that epoch. It is important to note that the models were not updated based on testing or null loss, only based on training loss.}
\label{fig:losses_by_epoch}
\end{figure}




\subsection{Meta- and hyper-parameter optimization}



\begin{figure}
    \centering
    \begin{tabular}{r}
    \includegraphics[width=.8\linewidth]{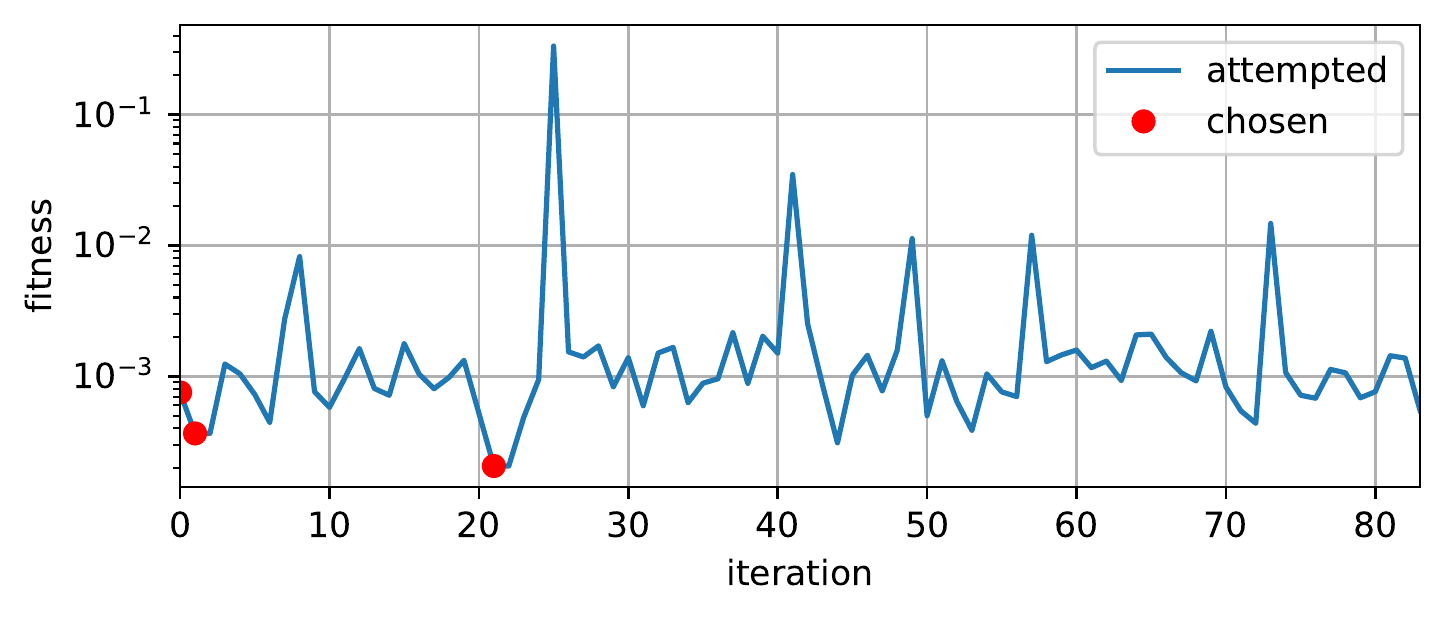} \\
    \includegraphics[width=.8\linewidth]{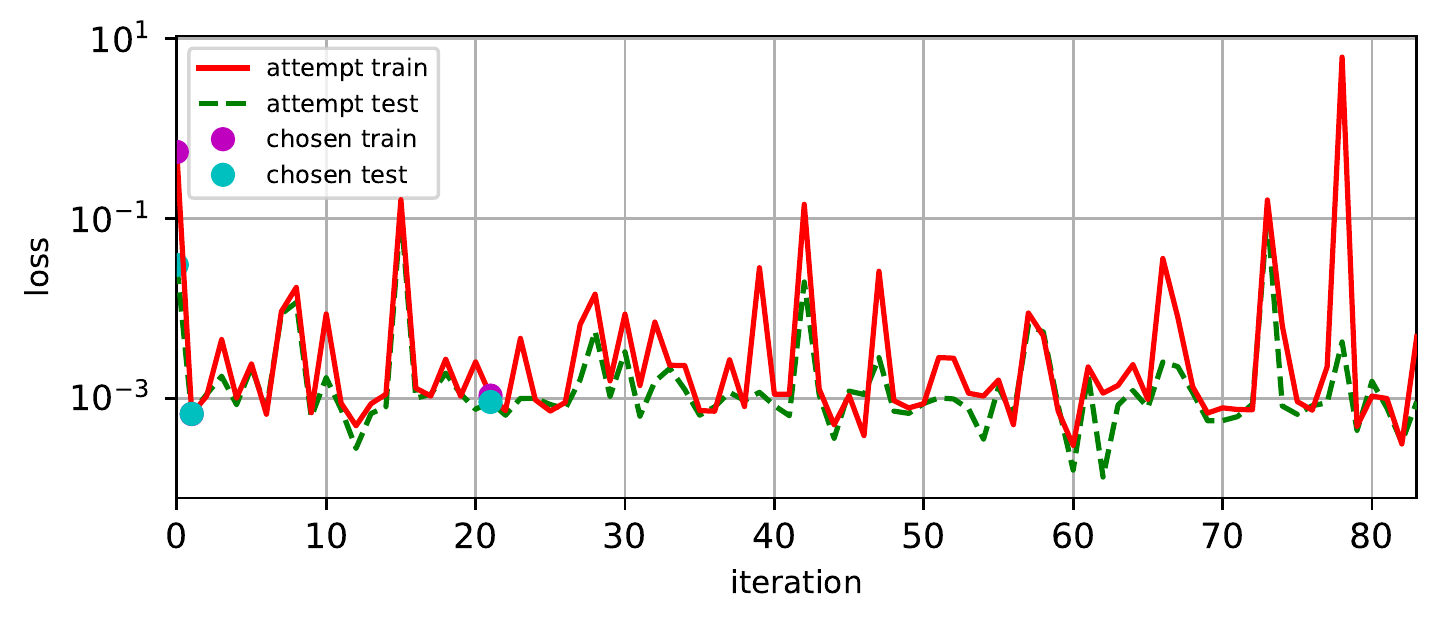}
    \end{tabular}
    \caption{Meta-parameter and hyper-parameter search results in the form of the (top) pattern search fitness and (bottom) final loss (averaged over dimensions) for each pattern search iteration. The zero iteration chosen meta-parameter set is the initial ``base point'' or randomly generated set of meta-parameters. 
    Subsequent chosen sets are exploratory points with the better fitness than the base point and their set. The ANN is trained for ten epochs with a given set of meta-parameters and hyper-parameters.}
    \label{fig:s10_fit}
\end{figure}

Figure \ref{fig:s10_fit} shows the pattern search fitness for each iteration. 
The first chosen value is the initial, randomly generated base point or set of meta-parameters.
The second chosen value is the set of meta-parameters that results in a ANN with the lowest fitness compared with the initial set of meta-parameters and the parent exploratory set of meta-parameter sets.
This chosen value is the new base point $X^{(\text{new})}$.
A pattern move creates a new temporary base point with a new temporary exploratory set. 
This sequence is repeated for the third chosen meta-parameter set.

Iterations after the third chosen meta-parameter did not result in a ANN with lower fitness.
The pattern search algorithm continued creating exploratory sets with changing mesh sizes probing the design space until the search limit was reached.
The maximum of number of exploratory sets without a change in base point is set at three for this run.
The maximum number of exploratory sets is set at three to reduce computation expenses as continuing the pattern search algorithm would be result in a large increase in computation time with nominal reduction in fitness.

{\renewcommand{\arraystretch}{1.3}
\setlength{\tabcolsep}{1ex}
\begin{table}[]
    \begin{center}
        \caption{Optimal hyper-parameters learning rate, optimizer parameters, and number of hidden layers. The first value of the activation function and features is the input layer.}
        \label{tab:hyper_optim_dim0}
    
        \begin{tabular}{lll}
        \toprule
            dimension          & hyper-parameter      & optimal values\\
            \midrule
            \multirow{3}{*}{0} & learning rate        & $3.76 \times 10^{-4}$ \\
                               & optimizer parameters & $\beta_1=0.900, \beta_2=0.996$\\
                               & hidden layers        & 8  \\ \cline{2-3}
                  
            \multirow{3}{*}{1} & learning rate        & $4.45\times 10^{-4}$\\
                               & optimizer parameters & $\beta_1=0.900, \beta_2=0.936$\\
                               & hidden layers        & 3\\
                               \cline{2-3}
            \multirow{3}{*}{2} & learning rate        & $5.06\times 10^{-3}$\\
                               & optimizer parameters & $\beta_1=0.950, \beta_2=0.936$\\
                               & hidden layers        & 2 \\
                               \cline{2-3}
            \multirow{3}{*}{3} & learning rate        & $9.25\times 10^{-3}$\\
                               & optimizer parameters & $\beta_1=0.900, \beta_2=0.992$\\
                               & hidden layers        & 5 \\
                               \bottomrule
        \end{tabular}
    \end{center}
\end{table}
}

{\renewcommand{\arraystretch}{1.3}
\setlength{\tabcolsep}{1ex}
\begin{table}[]
    \begin{center}
    \caption{Optimal hyper-parameters activation function and number of features for each ANN layer and projection/output dimension. Layer 0 is the input layer.}
    \label{tab:hyper_optim_activation_features}
    \begin{tabular}{lllllllllll}
    \toprule
    dim.\          & hyper-parameter     & \multicolumn{9}{c}{ANN layer}                                        \\
    & & 0 & 1 & 2 & 3 & 4 & 5 & 6 & 7 & 8 \\
    \midrule
    \multirow{2}{*}{0} & activation fun. & $\tanh$ & $\sigma$ & $\tanh$ & $\tanh$ & $\tanh$ & $\tanh$ & $\tanh$ & $\tanh$ & $\tanh$ \\
                       & features            & 760  & 88       & 8    & 8    & 8    & 8    & 8    & 8    & 8    \\ \cline{2-11}
    \multirow{2}{*}{1} & activation fun. &   $\tanh$   & $\sigma$    &   $\tanh$   &  $\tanh$    &      &      &      &      &      \\
                       & features            &   328   &     1368     &   8   &      &      &      &      &      &      \\ \cline{2-11}
    \multirow{2}{*}{2} & activation fun. &   $\tanh$   &     $S$     &   $\tanh$   &      &      &      &      &      &      \\
                       & features            &   784   &     1216     &   8  &      &      &      &      &      &      \\ \cline{2-11}
    \multirow{2}{*}{3} & activation fun. &   $\sigma$   &     $\tanh$     &  $\tanh$   &   $\tanh$   &    $\tanh$  &   $\tanh$   &      &      &      \\
                       & features            &   696   &     1168     &   8   &  8    &     8 &  8    &      &      &     \\
    \bottomrule
    \end{tabular}
    \end{center}
\end{table}
}

{\renewcommand{\arraystretch}{1.3}
\setlength{\tabcolsep}{1ex}
\begin{table}[]
    \begin{center}
    \caption{The optimal meta-parameters from the meta-parameter optimization.}
    \label{tab:optimal_metaparameters}
    \begin{tabular}{lll}
        \toprule
        & meta-parameter & optimal values \\
        \midrule
        \multirow{4}{*}{embeddings}
        & projection/output dimensions & $4$ \\
        & word em.\ model/input dimensions & $20$ \\
        & keyword weighting & $5$ \\
        & atom embedding method & PVDM \\ \cline{2-3}
        \multirow{3}{*}{ANN}
        & gradient-based optimizer & Adam \\
        & ANN architecture & FFNN \\
        & tuner optimizer & random search \\
        \bottomrule
    \end{tabular}
    \end{center}
\end{table}
}


\section{Discussion}


The results depend on a significant number of meta- and hyper-parameters, optimized according to the methods described above.
Beyond the results themselves, it is challenging to make general statements that are not potentially misleading; for instance, that technique X is better than technique Y.
However, with these qualifications and despite the risk of black swans,\footnote{See Hume and the ``problem of induction.''} we present comparisons among techniques despite the limited scope of evidence.











\subsection{Optimal meta-parameters and hyper-parameters}

The optimal hyper-parameters of \autoref{tab:hyper_optim_dim0} and \autoref{tab:hyper_optim_activation_features} and optimal meta-parameters of \autoref{tab:optimal_metaparameters} represent the ``best'' of each parameter found in this study.
Focusing on the meta-parameters of \autoref{tab:optimal_metaparameters}, the optimal number of projection/output dimensions, which is the dimension of the metastimuli, is four, a manageable number in terms of current haptic devices.
The optimal number of word embedding or model input dimension was $20$.
Keyword weighting optimized at five, which means scaling keyword embeddings derived directly from PIMS category names was effective (a weight of unity would imply no improvement); however, it is not so high as to imply that semantic content was irrelevant.
The distributed memory paragraph vector (PVDM) was the optimal atom embedding method, implying that it was superior to the bag-of-words and $\nabla$ methods, which is not particularly surprising given that it is a more advanced technique.

Regarding the gradient-based hyper-parameter optimizer technique, the Adam optimizer proved superior to SGD, AdaGrad, AdaDelta, AdaMax, and RMSprop.
The Adam optimizer is a stochastic gradient descent method developed by Kingma and Ba \cite{kingma2017adam}.
Surprisingly, the feed-forward neural network performed better than the recurrent neural network.
It should be noted, however, that the meta-parameter optimizer switched back-and-forth several times between the two architectures; we suspect further study may show the RNN to perform better.
The optimal hyper-parameter tuner optimizer was random search, outperforming the hyperband and Bayesian optimizers.

\subsection{Directions of the work}

The results presented here are promising: the ANNs can learn on relatively limited datasets and perform reasonably well at associating a meaningful (in terms of a user's PIMS) low-dimension real vector to atoms not previously seen.
However, much work remains before the hypothesis of the ``metastimulus bond effect'' \cite{picone2020} on human learning can be studied directly.
Before a study can be conducted, two significant results must be achieved:
\begin{enumerate}
	\item a haptic interface must be developed to apply the real vectors resulting from the work presented above (i.e.\ to apply metastimuli) and
	\item users (study participants) must be provided a software environment for creating their own PIMSs that can be processed by the PIMS filter software.
\end{enumerate}
Both directions are currently being pursued by the authors.

\section{Acknowledgements}

This work used the Extreme Science and Engineering Discovery Environment (XSEDE), which is supported by National Science Foundation grant number ACI-1548562, through the allocation for Shawn Duan (user: Dane Webb) \cite{xsede,ecss}. Specifically, it used the Bridges system (bridges-gpu.psc.xsede.org), which is supported by NSF award number ACI-1445606, at the Pittsburgh Supercomputing Center (PSC) \cite{nystrom2015}.
We thank TJ Olesky of PSC for their assistance with porting code over to Bridges.



\nocite{*} 
\bibliographystyle{splncs04}
\bibliography{../sources/hci_international_2021.bib}

\appendix

\section{Software repositories}

The software written to generate the results presented in this work is open-source and can be found in the permanent software repository of \cite{webb_dane_2021_4539755}.
The ongoing development of the software is hosted at the GitHub repository \cite{metastimuli_repo}.
Several sub-repositories are included: \cite{danewebb_class,rico_picone_2020_3633355,danewebb_tag}.



\end{document}